\pdfoutput=1

\documentclass[11pt]{article}

\usepackage{ACL2023}

\usepackage{times}
\usepackage{latexsym}

\usepackage[T1]{fontenc}

\usepackage[utf8]{inputenc}

\usepackage{microtype}

\usepackage{inconsolata}

\usepackage{amsmath,scalerel}
\usepackage{booktabs}
\usepackage{multirow}
\usepackage{xcolor}
\usepackage{colortbl}
\usepackage{tabularx}
\usepackage{subcaption}

\usepackage{algorithm,algorithmic}
\usepackage{listings}

\newcommand{\minus}{\scalebox{0.7}[1.0]{$-$}}
\definecolor{lightgreen}{rgb}{0.56, 0.93, 0.56}
\title{From Characters to Words: Hierarchical Pre-trained Language Model for Open-vocabulary Language Understanding}

\author{Li Sun{\normalfont \textsuperscript{1}$^\dagger$}, Florian Luisier{\normalfont \textsuperscript{2}}, Kayhan Batmanghelich{\normalfont \textsuperscript{1}}, Dinei Florencio{\normalfont \textsuperscript{2}}, {\bf Cha Zhang}\textsuperscript{2} \\ 
       \textsuperscript{1}Boston University  \hspace{0.5cm} \textsuperscript{2}Microsoft \\
$^{1}$\texttt{\{lisun,batman\}@bu.edu}, 
$^{2}$\texttt{\{flluisie,dinei,chazhang\}@microsoft.com}
}

\begin{document}
\maketitle
\def\thefootnote{$\dagger$}\footnotetext{Work done during internship at Microsoft.}
\def\thefootnote{}\footnotetext{PyTorch-like pseudocode is available in Appendix~\ref{app:algorithm}.}
\def\thefootnote{$\dagger$}
\begin{abstract}
Current state-of-the-art models for natural language understanding require a preprocessing step to convert raw text into discrete tokens. This process known as tokenization relies on a pre-built vocabulary of words or sub-word morphemes. This fixed vocabulary limits the model's robustness to spelling errors and its capacity to adapt to new domains. In this work, we introduce a novel open-vocabulary language model that adopts a hierarchical two-level approach: one at the word level and another at the sequence level. Concretely, we design an intra-word module that uses a shallow Transformer architecture to learn word representations from their characters, and a deep inter-word Transformer module that contextualizes each word representation by attending to the entire word sequence. Our model thus directly operates on character sequences with explicit awareness of word boundaries, but without biased sub-word or word-level vocabulary. 
Experiments on various downstream tasks show that our method outperforms strong baselines.
We also demonstrate that our hierarchical model is robust to textual corruption and domain shift.
\end{abstract}

\section{Introduction}
Pre-trained language models with Transformers have achieved breakthroughs in many natural language processing (NLP) tasks~\citep{devlin-etal-2019-bert,liu2019roberta}. One of the key advantages of Transformers over traditional feature-engineered NLP pipelines is that Transformers enable end-to-end training from vast amount of data to automatically learn the optimal language representation~\citep{mikolov2013distributed}. However, most recent language models still require a separate pre-processing stage known as tokenization. Tokenization is a process that splits raw text parts into a list of discrete tokens from a fixed vocabulary. This pre-defined vocabulary remains as an important bottleneck preventing truly end-to-end training of language models~\citep{tay2021charformer,islam2022vocabulary}.

\begin{table}
\centering
\renewcommand{\arraystretch}{1.5}
\resizebox{\columnwidth}{!}{\begin{tabular}{|c|c|c|}
\hline
&\textbf{Word}
&\textbf{BERT Tokens}\\
\hline
\multirow{2}*{\rotatebox[origin=c]{90}{\parbox[c]{1cm}{\small\centering \bf Spelling Errors}}}
&  changable (changeable)
&ch, \#\#anga, \#\#ble (change, \#\#able) \\
&outragous (outrageous)
&out, \#\#rag, \#\#ous (outrage, \#\#ous) \\
\hline
\multirow{2}*{\rotatebox[origin=c]{90}{\parbox[c]{1cm}{\small\centering \bf Domain Shift}}}
&reimbursement
&re, \#\#im, \#\#bur, \#\#se, \#\#ment \\
&invoice
&in, \#\#vo, \#\#ice \\
\hline
\end{tabular}}
\caption{Examples of sub-word tokenization of misspelled words (with their correct spellings in parentheses) and words from domain-specific corpus (e.g. business documents). Pre-trained sub-word tokenizer tends to over-segment words from these two categories, which produces less meaningful tokens.}
\label{tab:tokenization_example}
\end{table}


Based on the granularity of the basic token units, tokenization methods can be divided into three categories: character-based, subword-based and word-based. A word-based tokenizer segments sentence into smaller chunks of words. Due to language complexity and memory limit, a word-based vocabulary can not represent all possible words. Word-level tokenization thus frequently runs into the issue of \emph{out-of-vocabulary} words.
A character-based tokenizer simply splits the text into a sequence of its characters. It is flexible to encode arbitrary words, but character-level tokenization produces long sequences, which is undesirable as the computational cost of Transformers grows quadratically with the sequence length. To strike a good balance between time and space complexity, most state-of-the-art pre-trained language models thus adopt sub-word tokenization. 
Data-driven sub-word tokenizers~\citep{kudo2018sentencepiece,schuster2012japanese,kudo2018subword} are typically pre-trained on a general text corpus to learn a sub-word vocabulary based on the frequency of word fragments. 

Despite their popularity, sub-word tokenizers limit the robustness and generalizability of the language models built upon them. 
First, sub-word tokenizers are sensitive to small textual perturbations~\citep{xue-etal-2022-byt5}. While humans can still comprehend text with subtle misspellings and capitalization variants~\citep{rawlinson2007significance,davis2003psycholinguistic}, these perturbations can drastically change the tokenization results, potentially leading to a sub-optimal text representation.
Second, the sub-word vocabulary is pre-built and remains frozen during the language model pre-training and task-specific fine-tuning. Therefore, when adapting a pre-trained language model into a new language context (e.g. biomedical texts and business documents), the tokenizer is prone to excessive fragmentation of sub-word pieces~\citep{yasunaga2022linkbert,islam2022vocabulary}, as illustrated in Table~\ref{tab:tokenization_example}. While this issue could be partially remedied by further task-specific pre-training or by collecting more fine-tuning data, such mitigation would be costly and not always possible.

We aim to bring the best of both character-based and word-based models to address the challenges discussed above. To this end,
we propose a novel pre-trained language model with a hierarchical two-level architecture. At the word level, we split the text sequence by characters, and introduce an intra-word module that uses Transformers to learn a representation for each word in the sequence from the embeddings of their respective characters. At the sequence level, we introduce an inter-word module that contextualizes the embedding for every words in the text sequence. Our method does not require explicit sub-word or word-level vocabulary, and can thus be considered as an \emph{open-vocabulary} approach~\citep{mielke2021between}. By limiting the attention range to characters within the same word rather than the full sequence in the intra-word module, our model remains computationally efficient.

In order to validate our model, we comprehensively compare our method with various baseline methods, including the most popular sub-word based model BERT~\citep{devlin-etal-2019-bert}, some state-of-the-art character-based models~\citep{clark2022canine,boukkouri2020characterbert}, and an hybrid character/sub-word model~\citep{ma2020charbert}. Besides standard benchmarking, we also test the robustness of the various models in two ways: by introducing spelling noise into the validation set and by testing on cross-domain tasks.  

Our contributions can be summarized as follows:
\begin{itemize}
\item We introduce a novel open-vocabulary pre-trained language model with a hierarchical two-level architecture. 
Our method does not rely on pre-defined word or sub-word vocabulary.

\item We propose a novel adaptive and learnable aggregation method to summarize character-level features into word-level representations. An ablation study highlights its effectiveness.

\item We show that our method outperforms strong baselines on multiple benchmarking datasets, while being computationally efficient. 

\item We perform quantitative experiments and a case study to show that our model is robust to textual corruption and domain shift.

\end{itemize}

\section{Related Work}
\subsection{Word-level Models}
Word embedding methods including Word2vec~\citep{mikolov2013efficient} and GloVe~\citep{pennington-etal-2014-glove} led to many early NLP breakthroughs. These methods learn vector space representations of words from large-scale unlabeled corpora, and encode semantic relationships and meanings~\citep{goldberg2014word2vec}. 
In order to generalize to rare words, \citet{bhatia-etal-2016-morphological} proposed to use LSTM to learn word embedding from both morphological structure and word distribution.
While early methods only learned a context-independent word representation, ELMo~\citep{peters-etal-2018-deep} proposed to use a deep bidirectional language model to learn contextualized word representations.
In more recent studies, Transformer-XL~\citep{dai-etal-2019-transformer} enhanced the Transformer architecture with a recurrence mechanism to learn contextualized word embedding through language modeling.
Despite the recent progress, word-level models still face the out-of-vocabulary challenge for noisy text and non-canonical word forms~\citep{eisenstein-2013-bad}.

\subsection{Character-level Models}
Character-level language models emerged in the early years thanks to their simplicity and ability to better address \emph{out-of-vocabulary} words compared to word-level models~\citep{elman1990finding,graves2013generating,kalchbrenner2016neural}. While sub-word based approaches gained popularity in language modeling due to their superior performance, recent studies~\citep{choe2019bridging,xue-etal-2022-byt5} show that character/byte-level models can match the performance of their sub-word counterpart when provided with sufficient parameter capacity. In addition, character-level models have been shown to be more robust to text corruptions~\citep{tay2021charformer}, adversarial attacks, and domain shifts~\citep{aguilar2020char2subword}.

Character-level models also show promising results in multilingual settings. While sub-word or word tokenizers require a huge vocabulary to adequately cover various languages, multilingual character-based vocabulary can remain comprehensive and small. The text embedding layer does not eat up most of the model's parameter budget as in the multilingual BERT$_{Base}$ model for instance (up to 52\%). More parameters can then be dedicated to the Transformer layers in character-based approaches. Character-level models have also been shown to perform better on low-resource languages~\citep{islam2022vocabulary}.

An important drawback of character-level models is that they typically require more computations than sub-word and word-level models. This is because character-level tokenization produces longer token sequences compared to sub-word or word based approaches, and the computational and memory demands of the self-attention mechanism grow quadratically with the sequence length. In order to address this challenge, CANINE~\citep{clark-etal-2022-canine} leverages strided convolution to downsample the character sequence, while Charformer~\citep{tay2021charformer} uses average pooling. Although these methods improve the computational efficiency of character-level models, they require a predefined static downsampling rate. Such downsampling operation often breaks the boundary of basic linguistic units, including morphemes and words.

\subsection{Hybrid Models}
Vanilla character-level models do not explicitly extract word or sub-word morpheme representations, which might negatively impact their performance on word-level downstream tasks, including named-entity recognition and extractive question answering. In order to address this issue, there have been efforts to combine character-level and word/sub-word level approaches to build hybrid models.
These works propose to use information from character spelling to inform word representation.
For example, Flair~\citep{akbik-etal-2018-contextual} proposed to use the internal states of a pre-trained character language model to produce word-level embeddings. 
CharBERT~\citep{ma2020charbert} combined sub-word tokens and character tokens and fused their heterogeneous representations.
CharacterBERT~\citep{boukkouri2020characterbert} used a CNN to learn word-level representations from the embeddings of their characters, but still requires a word-level vocabulary for pre-training.
Char2Subword~\citep{aguilar2020char2subword} proposed a similar approach, where character embeddings are used to mimic pre-trained representation of sub-word tokens with Transformer encoder.


\section{Method}
\label{sec:method}
\begin{figure*}[t]
\centering
    \includegraphics[width = .9\textwidth]
    {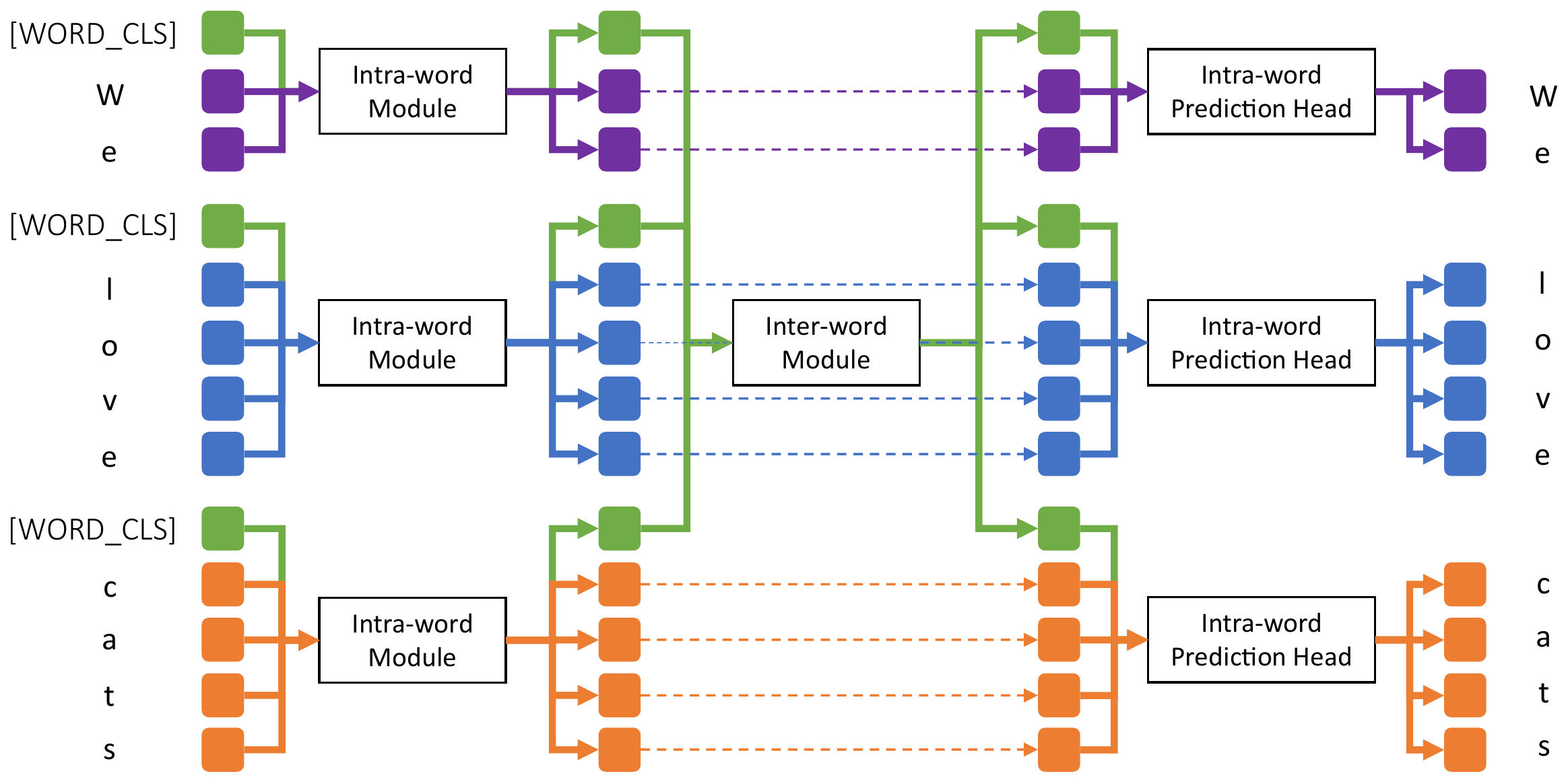}
    \caption{Overview of the proposed method. The intra-word module learns contextualized character embeddings by referring to characters from the same word. A $\texttt{[WORD\_CLS]}$ token is inserted at the beginning of each word to learn word-level representations. The inter-word module then learns contextualized word-level features by attending to all words in the sequence. Finally, the word-level and character-level embeddings are concatenated and fed to the intra-word prediction head for the pre-training task of masked character modeling. The prediction head is not used in downstream tasks.
    }
    \label{fig:main}
\end{figure*}

Most character-level Transformer encoder models are sub-optimal for two reasons: (1) Dense self-attention on long character sequence is computationally expensive; (2) They do not leverage word boundary, which is an important inductive bias in linguistics. To overcome these challenges, we propose to decompose dense character-level Transformer encoder into two parts: intra-word Transformer encoder and inter-word Transformer encoder.
Our hierarchical language model (HLM) adopts an hourglass structure~\citep{nawrot-etal-2022-hierarchical} and contains three main components: (1) an intra-word module that learns word embeddings from their characters; (2) an inter-word module which contextualizes the word representations by attending to all words in the input sequence;
(3) an intra-word prediction head for character-level pre-training. The overall architecture of our model is shown in Fig.~\ref{fig:main}. In the following sections, we discuss each component separately.

\subsection{Intra-word Module}
We aim to learn word-level representations from the embeddings of their characters. An ideal approach should be able to handle words of arbitrary lengths, attend to every character rather than a local window, and remain computationally efficient. Therefore, we choose a shallow (4 layers in our experiments) Transformer encoder to learn contextualized character embeddings, rather than a CNN or a LSTM used by previous methods~\citep{boukkouri2020characterbert,peters-etal-2018-deep}. Either average or max pooling~\citep{boukkouri2020characterbert,clark-etal-2022-canine} is commonly used to aggregate contextualized character embeddings and thus reduce the sequence length. However, such simple pooling tends to wash out strong signals from particular morphemes~\citep{FATHI2018229}. To address this challenge, we propose a novel adaptive and learnable aggregation method. 
Inspired by the approach of using the hidden state of the \texttt{[CLS]} token as the aggregate sequence-level representation,
we insert a special \texttt{[WORD\_CLS]} token at the beginning of every word. The embeddings of the \texttt{[WORD\_CLS]} tokens are then used as word-level representations. Formally, for the $i$-th word of $C_i$ characters in the sequence, we extract its word-level representation $\mathbf{r}^i$ as:
\begin{align*}
    \mathbf{h}^i &= f_{\theta}(\mathbf{e}^i_{0}\oplus \mathbf{e}^i_{1}\oplus \hdots \oplus \mathbf{e}^i_{C_i}) 
    \\
    \mathbf{r}^i &= \mathbf{h}^i_0,
  \end{align*}
  where $f_{\theta}$ is the intra-word Transformers that produces a contextualized representation $\mathbf{h}^i$ for each character of the $i$-th word, $\mathbf{e}^i_0$ is the embedding of the special \texttt{[WORD\_CLS]} token, $\mathbf{e}^i_c$ is the $c$-th character embedding of the $i$-th word, and $\oplus$ denotes concatenation along the sequence dimension.
  
In Sec.~\ref{sec:abl}, we conduct an ablation study to show that the proposed aggregation method outperforms the standard average or max pooling. By aggregating character-level tokens into word-level tokens, the token sequence length is greatly reduced for the subsequent inter-word module.

\subsection{Inter-word Module}
After obtaining word-level features, we apply an inter-word module consisting of deep transformer encoder layers to extract contextualized word-level representation by attending to all words in the sequences. Formally, the contextualized representation $\mathbf{w}^i$ of the $i$-th word of the sequence of $N$ words is given as:
\begin{align*}
    \mathbf{w}^i = f_{\phi}(\mathbf{r}^0\oplus\hdots\oplus\mathbf{r}^{N-1}),
\end{align*}
where $f_{\phi}$ denotes the inter-word Transformers.
  
We set the depth of the inter-word Transformer encoder stack to 12 in order to match the settings of BERT$_{Base}$~\citep{devlin-etal-2019-bert} and CANINE~\citep{clark-etal-2022-canine}.
The inter-word module contributes the most to the total model parameters. 

\subsection{Intra-word Prediction Head}
Since we adopt an open-vocabulary approach, we propose to use character-level masked language modeling as pre-training task. To restore the character-level token sequence, we concatenate the contextualized character representations from the intra-word module (the initial \texttt{[WORD\_CLS]} token is omitted) with the word-level features from the inter-word module along the sequence dimension. Finally, we apply a lightweight intra-word prediction head to get the posterior token probabilities. 
Formally, the prediction of the $C_i$ characters from the $i$-th word are given by:
\begin{align*}
    \mathbf{c}^i = f_{\sigma}(\mathbf{w}^i \oplus \mathbf{h}^i_1\oplus\hdots\oplus\mathbf{h}^i_{C_i}),
\end{align*}
where $f_{\sigma}$ is the intra-word prediction head, consisting of a single Transformer layer, a fully-connected layer and a Softmax layer.
Note that the intra-word prediction head is only used during pre-training for the masked character modeling task. 
During downstream fine-tuning, similar to CANINE, we concatenate initial word embedding $\mathbf{r}^i$ and contextualized word representation $\mathbf{w}^i$ along the feature dimension, and subsequently employ a small feed-forward network to integrate both low-level and high-level information for prediction. 

\subsection{Pre-training Task}
Following the practice of BERT, we pre-train our model on English Wikipedia and BookCorpus dataset (19G)~\citep{zhu2015aligning}. We pre-train the model for 3 epochs (3.9M steps with batch size set as 16) on a server with 8 NVIDIA Tesla V100 GPUs, and each epoch takes 137 hours. We adopt whole-word masked language modeling as pre-training task. In detail, we randomly select 15\% of words from the input sequence, and mask every characters in the selected word. 
We replace the character tokens in 80\% of the selected masked word with the \texttt{[MASK]} token. For 10\% of the selected masked words, we replace their characters with randomly selected characters drawn from our character vocabulary. 
The remaining 10\% words are unchanged.
The three main components of our model
are jointly trained in end-to-end fashion. 

\subsection{Implementation Details}
We use spaCy~\citep{honnibal2020spacy} to split sentences into words, which is rule-based using space, punctuation and special rules (\textit{e}.\textit{g}. splitting \emph{don't} into \emph{do} and \emph{n't}). We use a case-sensitive character vocabulary of size 1024, which consists of letters, digits and symbols. 
The maximum sequence length is set to 20 characters for the intra-word module and 512 words for the inter-word module.
A \texttt{[CLS]} and a \texttt{[SEP]} token are inserted at the beginning and end of each sequence respectively.
The hidden size is set to 768, the number of attention heads is set to 12, the feed-forward dimension in the Transformer encoder is set as 1536 and 3072 for intra-word and inter-word modules respectively.
We leverage relative position~\citep{he2021deberta} in our model, and we do not use token type embedding.
GELU~\citep{hendrycks2016gaussian} is used as activation function. Our model contains 125M parameters.
We use the AdamW optimizer~\citep{loshchilov2018decoupled} for model pre-training and fine-tuning. For the pre-training, the weight decay is set to $0.01$ and the number of warmup steps is set to 10,000. A linear learning rate decay schedule is used, starting at 5e-5.
The dropout rate is set to 0.1. More algorithm details can be found in Appendix~\ref{app:algorithm}.

\section{Experiments}

We evaluate the performance of our pre-trained model on a wide range of downstream tasks. 
We compare the performance of our pre-trained hierarchical language model (HLM) with various baseline methods, including the popular sub-word based BERT model, three recent byte/character-level models, as well as a hybrid model referred to as CharacterBERT. For BERT, we use the cased BERT$_{Base}$ model (108M parameters) to match our inter-word Transformers module setup. For CANINE, we adopt CANINE-C (132M) which also uses a character-level pre-training task. For CharacterBERT, we use the general version (105M) which is pre-trained on English Wikipedia and OpenWebText. For those baseline models, we use the pre-trained weights hosted on Huggingface\footnote{\url{https://huggingface.co/models}} or released by the authors. For Charformer (203M) and Byte-level T5 (200M), we use results of the base version from the original paper as pre-trained weight is not available.

\subsection{Evaluation on Standard Benchmarks}
\begin{table*}[t!]
        \begin{center}\small
        \begin{tabular}{l| cc | cc | ccc}
        \toprule
         &  \multicolumn{2}{c}{\bf SQuAD 1.1} & \multicolumn{2}{c}{\bf SQuAD 2.0 } & \bf  MRPC & \bf  QNLI & \bf MNLI (m/mm)\\
         Models&  EM &  F1 &   EM &   F1   &  Acc & Acc & Acc    \\ 
        \midrule
        BERT~\cite{devlin-etal-2019-bert}& 81.3 & 88.7  & 72.9 & 76.1 &  86.7  & 90.0 & 83.3/84.2\\
        Byte-level T5\footnotemark~\citep{xue-etal-2022-byt5}&-&-&-&-&87.3&88.7& 82.5/82.7\\
        Charformer~\citep{tay2021charformer}&-&-&-&-&87.3&89.0& 82.6/82.7\\
        CANINE~\cite{clark-etal-2022-canine} & 72.9 & 82.1 & 66.6 & 70.3 & 84.8 & 84.6 & 76.9/78.2 \\
        CharacterBERT~\cite{boukkouri2020characterbert} & 79.9 & 87.5 & 71.5 & 74.6 & 84.1 & 89.9 & 81.9/82.6 \\
        CharBERT~\cite{ma2020charbert} & 82.9 & 89.9 & 75.7 & 78.6 & 87.8 & \bf 91.7 & 82.9/83.1 \\
        HLM (Ours) &\bf 83.4 &\bf 90.4 & \bf 76.7 & \bf 79.9 & \bf 88.2 & 90.8 & \bf 84.4/84.3 \\
        \bottomrule
        \multicolumn{7}{p{.5\textwidth}}{- indicates not reported in the paper.}
        \end{tabular}
        \end{center}
        \vspace{-4mm}
        \caption{\label{result_clean} Experimental results on the validation set of question answering and text classification tasks. We report exact match (EM) and F1 scores for SQuAD, and accuracy for text classification tasks.}
         \end{table*}
In order to assess our model's performance on general domain, we evaluate our methods on standard English NLP benchmarks, including Stanford Question Answering Dataset (SQuAD) task~\citep{rajpurkar-etal-2016-squad,rajpurkar-etal-2018-know} and GLUE tasks~\citep{wang2018glue}. For the SQuAD task, we benchmark on both SQuAD 1.1 and 2.0 versions. SQuAD 1.1 dataset contains 100,000+ questions with associated context documents, and every question is answerable given the context. SQuAD 2.0 dataset contains an additional 50,000 unanswerable questions. 
We fine-tune the models for 2 epochs with a batch size of 16, and a learning rate of 3e-5.
The evaluation on the validation set is shown in Table~\ref{result_clean} (left). We use exact match (EM) and F1 scores as the two evaluation metrics. Our method outperforms all the baseline methods on both SQuAD versions.

We also benchmark our model on three text classification tasks from the widely adopted GLUE tasks~\citep{wang2018glue}, including MNLI~\citep{williams-etal-2018-broad}, MRPC~\citep{dolan2005automatically} and QNLI~\citep{rajpurkar-etal-2016-squad}. The MNLI dataset contains 393k training samples with textual entailment annotations. Given a sentence pair containing a premise and an hypothesis, the task is to predict whether the premise entails the hypothesis, contradicts the hypothesis, or neither. We conduct evaluation in both matched and mismatched settings. The MRPC dataset contains 3.7k of training sentence pairs, and the task is to predict whether the two sentences are semantically equivalent. The QNLI dataset contains 108k training samples of question-paragraph pairs, and the task is to predict whether the context sentence contains the answer to the question. We fine-tune the models on the datasets described above for 5 epochs, with a batch size of 16, and a learning rate of 2e-5. We use the accuracy as the evaluation metric. As shown in Table~\ref{result_clean}, our proposed method outperforms the baseline methods on all tasks.

In order to investigate the model's performance when the size is scaled up, we increase the size of our HLM to match BERT$_{Large}$ and benchmark the performance. The preliminary results can be found in Appendix~\ref{app:scaled_model}.

\footnotetext{Use results from \citet{tay2021charformer}}

\subsection{Robustness to Textual Corruptions}
\label{sec:perturbation}
Humans are prone to making spelling mistakes. For example, 10-15\% of web search queries contain misspellings~\citep{dalianis2002evaluating, cucerzan-brill-2004-spelling}.
In order to test our model's robustness to misspellings, we add synthetic noise to the fine-tuning and evaluation set of 
downstream tasks and re-evaluate all the models. 

Following the practice of \citet{xue-etal-2022-byt5}, we experiment with three types of noises: (1) \emph{Random drop}: We randomly delete 10\% of characters (spaces and punctuation are included) from the input sequence; (2) \emph{Random repeat}: We randomly select 20\% of characters, then append 1-3 repetitions (with equal probability) after the the selected original characters; (3) \emph{Random case}: We randomly set the case for each character (upper or lower) in the input sequence.

\begin{figure}
\centering
    \includegraphics[width = .48\textwidth]
    {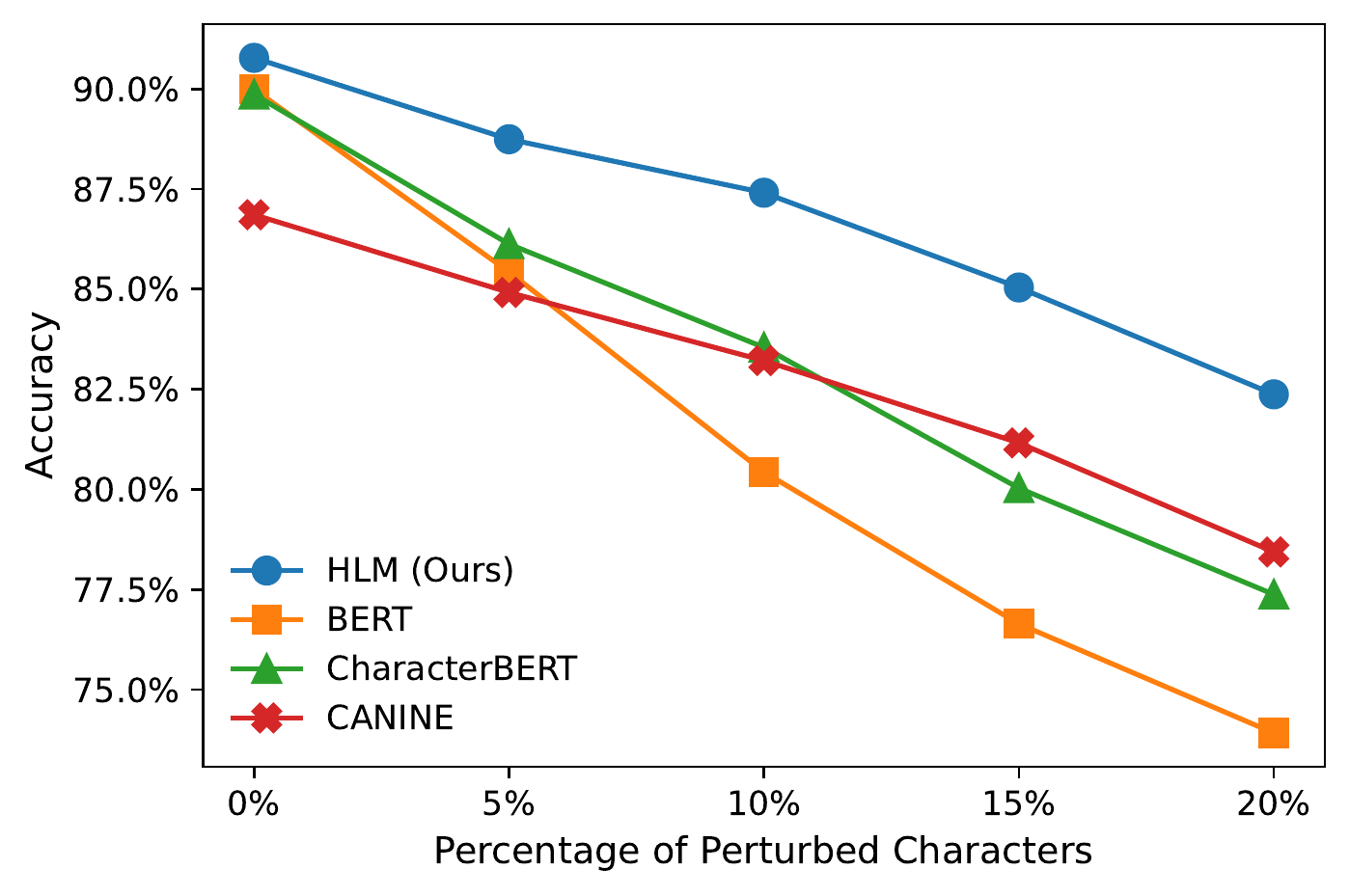}
    \caption{Evaluation results on QNLI under different amounts of noise. 
    With increasing amount of perturbation, our HLM consistently outperforms baseline methods, and the performance drop is smaller than BERT.}
    \label{fig:noise_trend}
\end{figure}

\begin{table*}[t!]
\centering
\footnotesize
\setlength\tabcolsep{4pt}
\begin{tabular}{llllll}
\toprule

& & \multicolumn{2}{c}{\bf MNLI(m/mm)} & \multicolumn{2}{c}{\bf SQuAD 2.0 } \\
& Model & Matched Acc & Mismatched Acc & \multicolumn{1}{c}{EM} & \multicolumn{1}{c}{F1}  \\
\midrule



\multirow{4}{*}{Random drop}
& BERT & 68.6(\minus14.7) & 70.0(\minus14.2) & 41.4(\minus30.6) & 47.6(\minus27.6) \\
& CANINE & 69.1(\minus7.8) & 69.9(\minus8.3) & 61.8(\minus4.7) & 64.8(\minus5.4)
\\
& CharacterBERT & 70.1(\minus11.8) & 71.1(\minus11.5) & 61.5(\minus9.9) & 64.4(\minus10.3)
\\
& HLM (Ours) & \textbf{74.7}(\minus9.7) & \textbf{75.2}(\minus8.9) & \textbf{69.3}(\minus7.4) & \textbf{72.2}(\minus7.6)\\

\midrule

\multirow{4}{*} {Random repeat}
& BERT & 72.3(\minus11.0) & 75.4(\minus8.8) & 35.2(\minus36.8) & 37.2(\minus38.0)\\
& CANINE & 76.3(\minus0.7) & 77.1(\minus1.2) & 65.3(\minus1.3) & 68.5(\minus1.7) \\
& CharacterBERT & 77.4(\minus4.4) & 78.2(\minus4.4) & 66.3(\minus5.1) & 69.4(\minus5.3)
\\
& HLM (Ours) & \textbf{83.1}(\minus1.3) & \textbf{82.8}(\minus1.4) & \textbf{75.8}(\minus0.9) & \textbf{78.9}(\minus1.0) \\
\midrule

\multirow{3}{*} {Random case}
& BERT & 71.2(\minus12.1) & 71.2(\minus13.0) & 35.9(\minus36.2) & 37.5(\minus37.7) \\
& CANINE & 76.7(\minus0.2) & 78.0(\minus0.2) & 66.1(\minus0.5) & 69.8(\minus0.5) \\
& HLM (Ours) & \textbf{83.5}(\minus0.9)&\textbf{83.5}(\minus0.7) & \textbf{76.3}(\minus0.4) & \textbf{79.3}(\minus0.5) \\

\bottomrule

\end{tabular}
\caption{Evaluation of models under various types of learnable noise. We apply the perturbations to both fine-tuning data and evaluation set.
We report the performance value and degradation compared to the standard evaluation in parentheses. Bold face indicates the best absolute performance. We do not report results for randomly switching case for CharacterBERT as it is an uncased model.}
\label{tab:noisy_tasks_learnable}
\end{table*}

 We perform the perturbation experiments on two representative downstream tasks: text classification on MNLI dataset and question answering on SQuAD 2.0. 
 For the MNLI dataset, we add noise to both premise and hypothesis sentences. 
 For the SQuAD 2.0 dataset, we only apply the perturbations to the question sentence, but not to the context paragraph, in order to avoid copying corrupted answer from the context for extractive QA models.
The evaluation results are shown in Table~\ref{tab:noisy_tasks_learnable}.
We found that BERT's performance significantly drops under perturbation, one explanation being that even subtle misspellings would greatly change the sub-word tokenization results. In comparison, character-level models including CANINE degrade less in the presence of noise. 
We also present the results for unseen perturbation setting in Appendix~\ref{app:robustness}.
Overall, our proposed HLM is robust to different kinds of perturbation and achieves the best performance.

In order to access the model's robustness to various magnitude of perturbations, we add different amounts of noise to the QNLI dataset and perform the evaluation. In practice, we randomly sample $5\%$, $10\%$, $15\%$, $20\%$ of characters for each example in the finetuning data and validation set. For each selected character, we either drop the character or repeat the character as mentioned above (equal probability). The accuracy on the validation set is shown in Fig.~\ref{fig:noise_trend}.

\subsection{Robustness to Domain Shift}
Most generic language models are pre-trained on web-crawled text corpora including Wikipedia and Common Crawl. But in real world deployments, models are often used in a different domain, an issue referred to as \emph{domain shift}. In order to evaluate the robustness to domain shift, we fine-tune and evaluate the pre-trained models on downstream tasks from specialized domains including biomedicine and social media. For the biomedical field, we perform the evaluation on the NCBI-disease dataset~\citep{crichton2017neural,gu2021domain}, which contains 7,287 sentences annotated with disease mentions from PubMed abstracts. The task is framed as a named entity recognition (NER) problem where the entities are the disease mentions.
We fine-tune the models for 20 epochs, with a batch size of 16, and a learning rate of 2e-5.
For the social media experiment, we leverage the W-NUT16 NER shared task~\citep{strauss-etal-2016-results}. This dataset contains 7,244 tweets annotated with 10 NER categories, including person, location, company and others. 
We fine-tune the models for 5 epochs.
The evaluation results on the test sets are shown in Table~\ref{tab:disbution_shift}. We use the F1 score as the evaluation metric. As observed, the proposed HLM outperforms the baseline methods, highlighting its higher robustness to domain shift. 

\begin{table}[t!]
\centering
\footnotesize
\setlength\tabcolsep{4pt}
\begin{tabular}{lcc}
\toprule

Model & NCBI-disease (F1) & W-NUT16 (F1)  \\
\midrule

BERT & 83.8 & 45.7 \\
CANINE & 75.2 & 32.0  \\
CharacterBERT & 84.7 & 34.0 \\
HLM (Ours) & \bf 86.4 & \bf 47.9 \\

\bottomrule

\end{tabular}
\caption{Evaluation results on cross-domain tasks. We report F1 score on the test set as the evaluation metric.}
\label{tab:disbution_shift}
\end{table}

\begin{table*}[t!]
\centering
\footnotesize
\setlength\tabcolsep{4pt}
\begin{tabularx}{\linewidth}{lcccccccccccccc}
\toprule
Text & \emph{Skin} & \emph{fragility} & \emph{in} & \emph{most} & \emph{cases} & \emph{is} & \emph{due} & \emph{to} & \emph{mutations} & \emph{in} & \emph{the} & \emph{gene} & \emph{encoding} & ...  \\
\midrule
BERT tokens & Skin & f, \#\#rag, \#\#ility & in & most & cases & is & due & to & mutations & in & the & gene & encoding &   \\
\midrule
BERT & \cellcolor{pink}O &\cellcolor{pink}O & O & O & O
& O & O & O & O & O
& O & O & O &  \\
HLM (Ours) & \cellcolor{lightgreen}B & \cellcolor{lightgreen}I & O & O & O
& O & O & O & O & O
& O & O & O &  \\
Label & B & I & O & O & O
& O & O & O & O & O
& O & O & O &  \\
\midrule
\end{tabularx}
\begin{tabularx}{\linewidth}{lcccccccccccccc}
Text & ... & \emph{a} & \emph{disease} & \emph{leading} & \emph{to} & \emph{aortic} & \emph{rupture} & \emph{in} & \emph{early} & \emph{adult} &\emph{life}  \\
\midrule
BERT tokens & & a & disease & leading & to & a, \#\#ort, \#\#ic & r, \#\#up, \#\#ture & in & early & adult &life   \\
\midrule
BERT & & O & O & O & O &\cellcolor{pink}O
&\cellcolor{pink}O & O & O & O & O\\
HLM (Ours) & & O & O & O & O & \cellcolor{lightgreen}B
& \cellcolor{lightgreen}I & O & O & O & O\\
Label & & O & O & O & O & B
& I & O & O & O & O\\
\bottomrule

\end{tabularx}
\caption{Case study of two examples from NCBI-disease NER task. The tagging schema for disease entities are beginning (B), inside (I), and outside (O). Pink/green colors indicate incorrect/correct predictions respectively.}
\label{tab:case_study}
\end{table*}

\begin{figure}[t!]
\centering
\begin{subfigure}{0.4\textwidth}
    \centering
    \includegraphics[width = \textwidth]
    {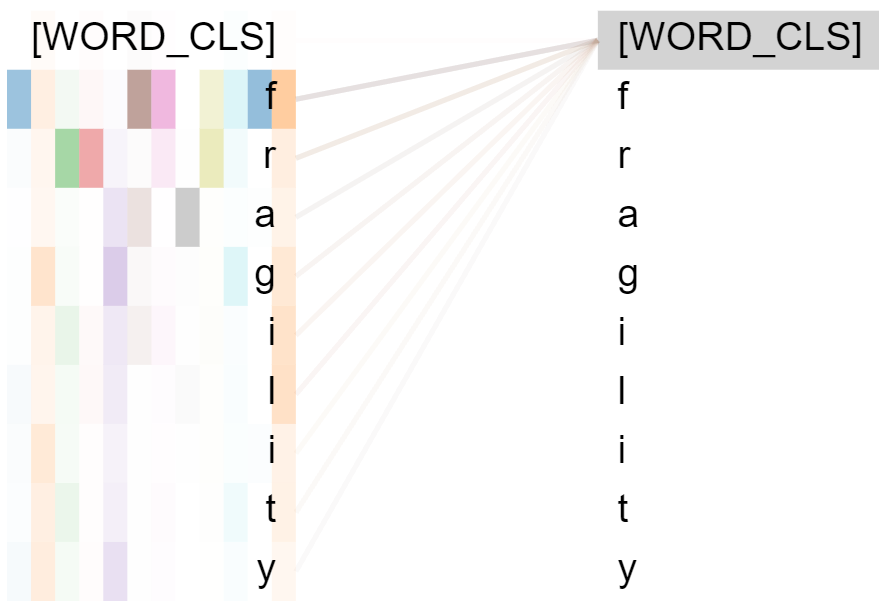}
\end{subfigure}
 \\
\hspace{1mm}
\begin{subfigure}{0.4\textwidth}
    \centering
    \includegraphics[width = \textwidth]
    {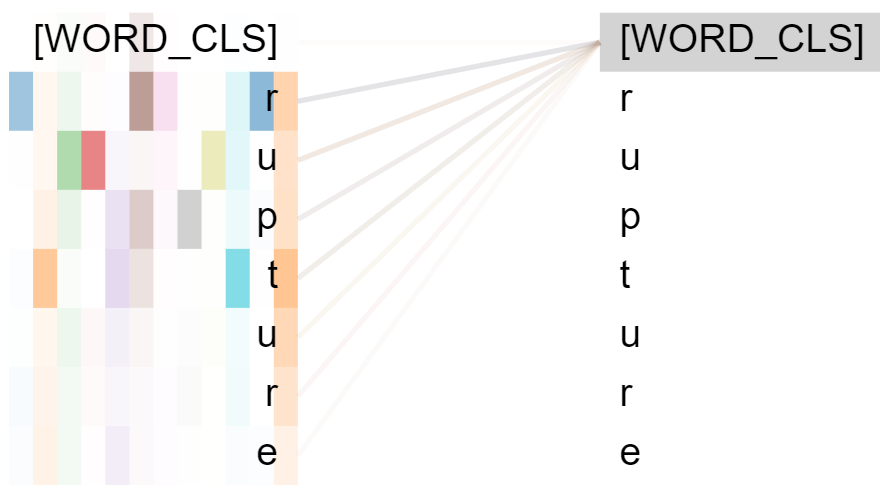}
\end{subfigure}
    \caption{Visualization of the attention patterns at the last layer of our intra-word module. Colored rectangles indicate the 12 attention heads. Color brightness and line weight reflect the attention scores. In the two examples, the $\texttt{[WORD\_CLS]}$ token is mainly attended by the stems of the words, \emph{fragil} and \emph{rupt}, respectively.
    }
    \label{fig:visualization_hlm}
\end{figure}

\textbf{Case study} In order to understand the performance gain of our model over sub-word based BERT on cross-domain tasks, we look into the cases where BERT makes incorrect predictions. We found that many of these cases contain excessively fragmented words. Table~\ref{tab:case_study} shows two examples from the NCBI-disease NER task. The word \emph{fragility} in case 1 is segmented into \emph{f, \#\#rag, \#\#ility}, and the word \emph{rupture} in case 2 is segmented into \emph{r, \#\#up, \#\#ture}. We think these tokenization results are sub-optimal as they break word morphemes, which possibly explains BERT's mispredictions. In comparison, we use BertViz~\citep{vig-2019-multiscale} to visualize the behavior of our HLM model. Specifically, we visualize the attention patterns of the $\texttt{[WORD\_CLS]}$ token of the last Transformer layer of our intra-word module. As shown in Fig.~\ref{fig:visualization_hlm}, the $\texttt{[WORD\_CLS]}$ token for the word \emph{fragility} and \emph{rupture} are primarily attended by the character string \emph{fragil} and \emph{rupt} respectively, which are the stems of the words.

\subsection{Ablation Study}
\label{sec:abl}
In this section, we perform an ablation study to compare the effect of different word-level aggregation methods. Specifically, we replace the proposed special token learning-based aggregation with standard aggregation methods such as average pooling and max pooling. We did not implement the strided convolution proposed in CANINE as it can not handle the variable word lengths. We report the validation accuracy on MRPC and the test F1 score on NCBI-disease in Table~\ref{tab:ab}. Our learned aggregation outperforms the standard pooling strategies. Note that average and max pooling are usually performed on a fixed-length window of characters in previous studies~\citep{tay2021charformer}, not adaptively at the word-level as in our ablation study.
\begin{table}
\centering
\footnotesize
\setlength\tabcolsep{4pt}
\begin{tabular}{l|ccc}
\toprule

Dataset & Average pooling & Max pooling & Ours  \\
\midrule
MRPC (Acc) & 82.1 & 83.6 & \bf 86.0 \\
NCBI-disease (F1) & 85.3 & 85.9 & \bf 86.6\\
\bottomrule

\end{tabular}
\caption{Word-level aggregation comparisons. All models are pre-trained for 1.5 epochs.}
\label{tab:ab}
\end{table}

\begin{table}[t!]
\centering
\footnotesize
\setlength\tabcolsep{4pt}
\begin{tabular}{lc}
\toprule

Model & Throughput (sample/sec)  \\
\midrule

BERT & \bf 93.8 \\
CANINE & 44.3  \\
CharacterBERT & 78.4 \\
HLM (Ours) & 90.3 \\

\bottomrule

\end{tabular}
\caption{Evaluation results on computational efficiency.}
\label{tab:throughput}
\end{table}

\subsection{Computational Efficiency}
In this section, we benchmark the computational efficiency of the proposed model. Specifically, we measure the inference throughput (number of processed samples per second) on the test set of the MRPC dataset, a sub-task of the GLUE benchmark. We evaluate different models on the same server with one NVIDIA Tesla V100 GPU. The batch size is set to 32 and we use single precision. The evaluation results are shown in Table~\ref{tab:throughput}. While BERT is the most computationally efficient, our HLM also performs competitively, the performance gap being smaller compared to other character-level baseline models. We speculate that this performance gain comes from our hierarchical architecture. By aggregating character tokens into word-level tokens, the sequence length is drastically reduced for the inter-word module which has the deepest Transformer stack. We provide more analysis on the computational complexity in Appendix~\ref{app:complexity}.

\section{Conclusion}
In this work, we propose a novel hierarchical language model for open-vocabulary language understanding. 
Our method does not rely on explicit sub-word or word vocabulary. We demonstrate that our HLM model outperforms baseline methods on standard benchmarks, and highlight its robustness to spelling errors and domain shifts. In future work, we will expand our language support and explore incorporating a decoder for generative tasks. 

\section*{Limitations}
This work has two main limitations. First, we only consider baseline models with similar amount of parameters, and pre-trained on similar scale of text corpus for comparison. While we are aware of recent models including T5~\citep{raffel2020exploring} and PaLM~\citep{chowdhery2022palm}, they either use huge corpus like C4 (745GB text) for pre-training or  contain significantly more parameters than ours. In the future, we will try to find additional computational resources to scale up our model and pre-train on larger text corpus. Second, we leverage spaCy to segment sentences into words, which is rule-based using spaces, punctuations and other rules. This approach works well on English and many other common languages such as French, German and Spanish. But for a few languages that do not use spaces to split words (e.g. Chinese and Japanese), it will be challenging to retrieve word boundaries. To address this issue, we consider either falling back to character splitting for these languages (similar to multilingual BERT) or employing a more sophisticated word boundary detector in future work. 

\section*{Acknowledgements}

This work was partially supported by NIH Award Number
1R01HL141813-01 and the Pennsylvania Department of
Health. We are grateful for the computational resources
provided by Pittsburgh Super Computing grant number TGASC170024.
\bibliography{anthology,custom}
\bibliographystyle{acl_natbib}
\newpage
\appendix

\section{Appendix}
\label{sec:appendix}
\subsection{Analysis on Computational Complexity}
\label{app:complexity}
Let $N$ denotes the character length of input sequence. Without loss of generality, we assume the words in the sequence are of the same length $M$. The multi-head self-attention module is the major component of Transformer. While it provides global receptive field, the computational cost and memory footprint grow quadratically with input sequence length~\citep{zeng2021you}. Therefore, for a vanilla character-based Transformers with dense self-attention, the computational and space complexity is
$O(N^2)$.

For our proposed HLM, the input sequence is still at the character level. But we sparsity the dense self-attention by introducing a hierarchical architecture. For the intra-word module, each character token only attends to characters from the same word. Since there are $\frac{N}{M}$ words in the sequence, the computational and space complexity of the intra-word module is
\begin{equation}
    O\left(\frac{N}{M}\cdot M^2\right) = O(NM)
\label{eq:intra}
\end{equation}

For the inter-word module, since it only operates on word-level tokens, the computational and space complexity is 
\begin{equation}
    O\left(\frac{N^2}{M^2}\right)
\label{eq:inter}
\end{equation}

Since typically $N\gg M$, and we have a shallow intra-word module and a deeper inter-word module, Eq.~\ref{eq:inter} dominates the computational and space complexity of the full model, which is significantly lower than the vanilla character-level model.

In comparison to sub-word based models like BERT,  our inter-word module operates on word-level token sequence, which is always equal or shorter than sub-word level token sequence. Therefore, despite our model has an extra intra-word module, we empirically observe in Table~\ref{tab:throughput} that our HLM is competitive in terms of computational efficiency comparing to sub-word based models. 


\subsection{Preliminary Evaluation of Scaled Model}
\label{app:scaled_model}
In this section, we scale up our model size and benchmark the performance. 
In order to match BERT$_{Large}$, we set the number of layers in the inter-word Transformer encoder to 24 and the feed-forward dimension of Transformer encoder is set as 2048 and 4096  for intra-word and inter-word modules respectively. We set the number of attention heads as 16 and the hidden size as 1024. The batch size is set as 128. Other hyperparameters are set as the same as  HLM$_{Base}$, described in Section~\ref{sec:method}. Due to limited access to computational resources, we could only pre-train the model for 370k steps at the camera-ready deadline. In comparison, BERT$_{Large}$ was pretrained for 1M steps with a batch size of 256. Therefore, our computational budget is about 1/6 of BERT’s. We benchmark our model's performance on SQuAD dataset. The evaluation on the validation set is shown in Table~\ref{app:result_clean}. We use exact match (EM) and F1 scores as the two evaluation metrics. 
Our model performs competitively compared with BERT$_{Large}$, despite that our HLM$_{Large}$ has significantly less computational budget for pre-training. 

\begin{table}
        \begin{center}\small
        \begin{tabular}{l| cc|cc }
        \toprule
         &   \multicolumn{2}{c}{\bf SQuAD 1.1}
         &   \multicolumn{2}{c}{\bf SQuAD 2.0 } \\
         Models&   EM &  F1& EM &  F1 \\ 
        \midrule
        BERT$_{Large}$& 84.1&90.9&78.7&81.9\\
        HLM$_{Large}$ (370k steps) &83.4&90.2&78.2&81.3 \\
        \bottomrule
        \end{tabular}
        \end{center}
        \vspace{-4mm}
        \caption{\label{app:result_clean} Experimental results on the validation set of question answering tasks. We report exact match (EM) and F1 scores for SQuAD.}
         \end{table}

\subsection{Algorithm Details}
\label{app:algorithm}
In this section, we provide algorithm details for our input pre-processing and model algorithm. Our pre-processing consists the following steps. First, we split each sentence into a list of words. 
Next, we map characters to codepoint indexes using a character-level vocabulary, and insert $\texttt{[WORD\_CLS]}$ token at the start of
each word. 
Next, we insert a $\texttt{[CLS]}$ token at the start, and a $\texttt{[SEP]}$ token at the end for each sequence.
Then we truncate the token sequence based on both character-level (20 characters for each word) and word-level (512 words per sentence) limits. Next, we compute the maximum number of characters for words in the batch, and pad all words to this length.
We also determine the maximum number of words in the sequence batch, and pad all sequences to this length. The pre-processed batch can then be represented as a matrix of shape $\texttt{[batch\_size, max\_num\_word, max\_num\_char]}$. Our unique representation of text batch enables us to efficiently switch between performing intra-word self-attention and inter-word self-attention by simply reshaping, which is shown in Algorithm~\ref{alg:code}.

We provide pseudocode for pre-training of our HLM in Algorithm~\ref{alg:code}. For better readability, we omit implementation details including utilizing attention mask which avoids performing attention on the $\texttt{[PAD]}$ tokens and handling for padding words. We recommend padding the input matrix to multiples of 8 for better acceleration on GPU. We also found a residual connection between initial word embedding $\mathbf{r}^i$ and contextualized word embedding $\mathbf{w}^i$ improves the performance in a subsequent study.

\begin{algorithm*}
\caption{Pseudocode for HLM, PyTorch-like}
\label{alg:code}
\definecolor{codeblue}{rgb}{0.25,0.5,0.5}
\definecolor{codekw}{rgb}{0.85, 0.18, 0.50}
\lstset{
  backgroundcolor=\color{white},
  basicstyle=\fontsize{9.5pt}{9.5pt}\ttfamily\selectfont,
  columns=fullflexible,
  breaklines=true,
  captionpos=b,
  commentstyle=\fontsize{9.5pt}{9.5pt}\color{codeblue},
  keywordstyle=\fontsize{9.5pt}{9.5pt}\color{codekw},
}
\begin{lstlisting}[language=python]
# embeddings: character-level embedding lookup table 
# intra_word_encoder: Intra-word Transformer encoder
# inter_word_encoder: Inter-word Transformer encoder
# intra_word_head: Intra-word prediction head

for input_ids, labels in loader:  # load a minibatch with n samples
    input_embeds = embeddings(input_ids)
    batch_size, num_word, num_char, hidden_size = input_embeds.shape
    
    # reshape to let Transformers attend to intra-word tokens rather than full sequence
    input_embeds = input_embeds.reshape((batch_size*num_word, num_char, hidden_size))
    initial_embeds = intra_word_encoder(input_embeds)
    
    # extract embedding for [WORD_CLS] token, which is always at the beginning of each word
    word_embeds = initial_embeds[:,0,:]
    
    # reshape and extract contextualized inter-word representation
    word_embeds = word_embeds.reshape((batch_size, num_word, hidden_size))
    word_embeds = inter_word_encoder(word_embeds)
    word_embeds = word_embeds.reshape((batch_size*num_word, 1, hidden_size))
    
    # concatenate to restore the character-level token sequence
    char_embeds = concatenate([word_embeds, initial_embeds[:,1:,:]], axis=1)
    char_logits = intra_word_head(char_embeds)
    char_logits = char_logits.reshape((batch_size, num_word, num_char, -1))
    
    loss = CrossEntropyLoss(char_logits, labels) # masked character modeling loss
    loss.backward()  # back-propagate
    # AdamW update
    update(embeddings, intra_word_encoder, inter_word_encoder, intra_word_head)  

\end{lstlisting}
\end{algorithm*}

\subsection{Robustness to Unseen Perturbations}
\label{app:robustness}
In this section, we benchmark the model's robustness to \emph{unseen noise}. Specifically, we only add noise to the evaluation set, while using the original fine-tuning data.
We experiment with three types of perturbation as introduced in Section~\ref{sec:perturbation}. The results are shown in Table~\ref{tab:noisy_tasks_unseen}. In all three scenarios, our proposed HLM outperforms baseline methods, showing better robustness.
\begin{table*}
\centering
\footnotesize
\setlength\tabcolsep{4pt}
\begin{tabular}{llllll}
\toprule

& & \multicolumn{2}{c}{\bf MNLI(m/mm)} & \multicolumn{2}{c}{\bf SQuAD 2.0 } \\
& Model & Matched Acc & Mismatched Acc & \multicolumn{1}{c}{EM} & \multicolumn{1}{c}{F1}  \\
\midrule



\multirow{4}{*}{Random drop}
& BERT & 57.5(\minus25.8) & 57.9(\minus26.3) & 53.1(\minus19.0) & 55.6(\minus19.6) \\
& CANINE & 57.7(\minus19.2) & 58.2(\minus20.1) & 57.1(\minus9.5) & 59.0(\minus11.3)
\\
& CharacterBERT & 55.9(\minus26.0) & 56.0(\minus26.6) & 52.0(\minus19.5) & 55.1(\minus19.5)
\\
& HLM (Ours) & \textbf{59.7}(\minus24.7) & \textbf{61.0}(\minus23.2) & \textbf{58.3}(\minus18.4) & \textbf{60.3}(\minus19.5)\\

\midrule

\multirow{4}{*} {Random repeat}
& BERT & 52.4(\minus30.9) & 53.5(\minus30.7) &
51.4(\minus20.7) & 52.7(\minus22.5)\\
& CANINE & 56.2(\minus20.7) & 57.4(\minus20.8) &
53.8(\minus12.8) & 56.1(\minus14.2) \\
& CharacterBERT & 54.5(\minus27.4) & 55.2(\minus27.4) 
& 49.4(\minus22.1) & 52.6(\minus22.0)
\\
& HLM (Ours) & \textbf{58.5}(\minus25.9) & \textbf{58.3}(\minus25.9) 
& \textbf{57.7}(\minus19.1) & \textbf{59.2}(\minus20.7) \\
\midrule

\multirow{3}{*} {Random case}
& BERT & 43.8(\minus39.5) & 44.1(\minus40.2) &
48.1(\minus23.9) & 48.4(\minus26.8) \\
& CANINE & 72.7(\minus4.2) & 73.2(\minus5.1) &
65.3(\minus1.3) & 68.6(\minus1.6) \\
& HLM (Ours) & \textbf{73.5}(\minus10.9)&\textbf{74.5}(\minus9.6) & \textbf{70.2}(\minus6.5) & \textbf{73.1}(\minus6.8) \\

\bottomrule

\end{tabular}
\caption{Evaluation of the models under various types of unseen noise. The perturbations are only applied to the evaluation sets, while the fine-tuning data is left untouched.
We report the performance value and degradation compared to the standard evaluation (no perturbation) in parentheses. Bold face indicates the best absolute performance. We do not report results for randomly switching case for CharacterBERT as it is an uncased model.}
\label{tab:noisy_tasks_unseen}
\end{table*}

\end{document}